\documentclass[manuscript,screen]{acmart}

\usepackage{subfigure}
\usepackage{multirow}
\usepackage{multirow}
\usepackage{hhline} 
\usepackage{tabularx}
\usepackage{makecell}
\usepackage[ruled]{algorithm2e}

\AtBeginDocument{%
  \providecommand\BibTeX{{%
    \normalfont B\kern-0.5em{\scshape i\kern-0.25em b}\kern-0.8em\TeX}}}

\begin{document}

\title{Depth image denoising using nuclear norm and learning graph model}

\author{Chenggang~Yan}
\email{cgyan@hdu.edu.cn}
\affiliation{%
  \institution{Hangzhou Dianzi University}
  \city{Hangzhou}
  \state{China}
}
\author{Zhisheng~Li}
\email{zhisheng.li.go0411@outlook.com}
\affiliation{%
  \institution{Hangzhou Dianzi University}
  \city{Hangzhou}
  \state{China}
}
\author{Yongbing~Zhang}
\email{zhang.yongbing@sz.tsinghua.edu.cn}
\affiliation{%
  \institution{Tsinghua Shenzhen International Graduate School, Tsinghua University}
  \city{Shenzhen}
  \state{China}
}
\author{Yutao~Liu}
\email{liuyutao2008@gmail.com}
\affiliation{%
  \institution{Tsinghua Shenzhen International Graduate School, Tsinghua University}
  \city{Shenzhen}
  \state{China}
}
\author{Xiangyang~Ji}
\email{xyji@tsinghua.edu.cn}
\affiliation{%
  \institution{Tsinghua University}
  \city{Beijing}
  \state{China}
}
\author{Yongdong~Zhang}
\email{zhyd73@ustc.edu.cn}
\affiliation{%
  \institution{University of Science and Technology of China}
  \city{Hefei}
  \state{China}
}

\thanks{This work was supported by National Nature Science Foundation of China (61931008, 61671196, 61701149, 61801157, 61971268, 61901145, 61901150, 61972123, 61922048), by the National Natural Science Major Foundation of Research Instrumentation of PR China under Grants 61427808, Zhejiang Province Nature Science Foundation of China (LR17F030006, Q19F010030), 111 Project, No. D17019, the China Postdoctoral Science Foundation under Grant 2019M650686. (Corresponding author: Yongbing Zhang and Yutao Liu.)}


\begin{abstract}
The depth images denoising are increasingly becoming the hot research topic nowadays because they reflect the three-dimensional (3D) scene and can be applied in various fields of computer vision. But the depth images obtained from depth camera usually contain stains such as noise, which greatly impairs the performance of depth related applications. In this paper, considering that group-based image restoration methods are more effective in gathering the similarity among patches, a group based nuclear norm and learning graph (GNNLG) model was proposed. For each patch, we find and group the most similar patches within a searching window. The intrinsic low-rank property of the grouped patches is exploited in our model. In addition, we studied the manifold learning method and devised an effective optimized learning strategy to obtain the graph Laplacian matrix, which reflects the topological structure of image, to further impose the smoothing priors to the denoised depth image. To achieve fast speed and high convergence, the alternating direction method of multipliers (ADMM) is proposed to solve our GNNLG. The experimental results show that the proposed method is superior to other current state-of-the-art denoising methods in both subjective and objective criterion.
\end{abstract}

\keywords{learning graph model, low-rank, nonlocal self-similarity, ADMM.}

\maketitle

\section{Introduction}
The depth image, which is able to describe the distance between the 3D objects and the camera plane, can be acquired by depth cameras in our daily life \cite{yan44}. Although the depth images can be applied in image segmentation, object detection, target tracking, gesture recognition and other applications in computer vision \cite{tian2009view}\cite{yan33}\cite{yan22}\cite{yan11}\cite{yan55}, the depth images obtained from depth camera directly are usually corrupted by noises due to imperfect depth sensing technology. To acquire more clean information from degraded images, many researchers embark on the area of depth image denoising\cite{chen2017depth}\cite{hu2013depth}. The general description of denoising problem is derived from the expression: $y=x+n$, where $n$ is an additive white Gaussian noise or other kinds of noise\cite{Xie2015Joint}\cite{zhang2017beyond}. Our goal is to recover the original image with vector form $x$ from the noisy version $y$ as efficiently as possible. Many image denoising methods have been proposed in the past few decades.
\par One of the most popular image denoising algorithms is wavelet transform denoising \cite{buades2005review}\cite{shensa1992discrete}, which is a transform-domain filtering method. Besides the wavelet transform, Fourier transform and mean filter are also the common methods in image denoising. But the conventional methods don't take the internal relations of image into account, which usually lead to under-fitting or over-fitting results. To address this problem, the image priors reflecting the image intrinsic properties are usually considered in image denoising. Total variation is the earliest model that introduces the image prior \cite{wang2017non}, which considers the image prior based on the gradient of the image. The early solution can be formulated as a minimization problem:
\begin{equation}
  \mathop{\arg\min}_{x}\frac{1}{2}{\left \| {y-x}\right \|_2^2+\lambda\Psi(x)},
\end{equation}
where $\frac{1}{2}\left \| {y-x}\right \|_2^2$ is the $\l_2$ data-fidelity term, $\Psi(x)$ is a regularization fuction of image $x$ that reflects the prior knowledge, $\lambda$ is the tune parameter of $\Psi(x)$. Besides, low rank prior derived from the matrix completion is also utilized in image processing \cite{ji2010robust}\cite{Liu2018Image}, which achieved great performance in image denoising.
\par In the last few years, the dictionary learning has been emerging in the image restoration \cite{Elad2006Image, Zhu2014Fast, Wang2017Sparse}, which reflects the image sparsity property. Although the sparse dictionary representation is able to exploit the underlying structure of data matrix, and consequently achieves good performance, it is a non-convex problem and often needs high computational complexity \cite{WangMarginalized}. Besides the sparse property, the nonlocal property that is derived from the nonlocal similar patches correlating with each other within an image also draws much attention in image reconstruction \cite{Buades2005A}\cite{, Xie2015Joint}. As a typical example, nonlocal means (NLM) filter is an effective denoising method that takes advantages of the nonlocal self-similarity \cite{Dong2017Color}, which can restore the unknown pixels with high accuracy. In addition, manifold study is also a popular technique in image restoration \cite{Huang2013Self}. In manifold study, the graph is one of the most popular expressions, which can capture the intrinsic structure of the data matrix. Meanwhile, the graph model also reflects the nonlocal self-similarity in image processing. The manifold Laplacian matrix encapsulates the graph topology, which plays a crucial role in the graph signal processing \cite{Yankelevsky2017Dual}\cite{Deng2014Weakly}. Generally, the graph model can be denoted by the corresponding manifold Laplacian, and the construction of Laplacian matrix is based on the internal similarity of data matrix.
\par Inspired by the work in \cite{hu2013depth, li2017cfmda, kalofolias2014matrix},
this paper proposes a group based nuclear norm and learning graph (GNNLG) to solve the denoising problem, which combines the low rank and self-similarity property of the depth image. The alternating direction method of multipliers (ADMM) is employed to solve our proposed method. The experimental results show that our proposed GNNLG outperforms other state-of-the-art methods. The major contributions of our paper are as follows:
\newline 1) The newly graph Laplacian matrix strategy is provided in our paper, which can enforce the smoothing effect of the recovered depth image. The learning strategy is an optimization formulation, which includes the basic graph construction and the Laplacian regularization.
\newline 2) We exploit the intrinsic low rank and nonlocal self-similarity properties of depth image and adopt an unified framework to solve the problem. Meanwhile, it is notable that the combination of nuclear norm and learning graph model is firstly applied in image denoising area.
\newline 3) We propose a fast threshold algorithm rather than conventional soft threshold to tackle the nuclear norm subproblem. Besides the inner iteration using ADMM algorithm, we propose an overall iterative regularization scheme to optimize the results further.
\par The remainder of this paper is organized as follows. Section \uppercase\expandafter{\romannumeral2} gives the overview of the related work. The detailed solution and related optimization of our algorithm are shown in Section \uppercase\expandafter{\romannumeral3}. The experimental results and analysis are provided in Section \uppercase\expandafter{\romannumeral4}. Finally, Section \uppercase\expandafter{\romannumeral5} concludes our work and gives our future work.

\section{Related work}
In the last decade, image denoising has attracted more and more research interests and got an unprecedented development. Early works in image denoising mainly consist of conventional spatial filtering and transform-domain filtering methods. Zhang et al. \cite{zhang2013two} used the basic mean filter method for denoising, which is one of the most famous classic spatial filtering methods. Tomassi et al. \cite{tomassi2015wavelet} proposed the classic wavelet transform method and achieved great denoising results. In addition, wavelet transform was expanded by many researchers, which is proved to be the most classic transform-domain filtering method in image denoising. But conventional spatial and transform-domain methods ignore the inherent property of images, which results in a blurry result. Recently, the sparsity, low rank and self-similarity property of the image attracted much attention of researchers.
\par Ivan \cite{selesnick2017total} proposed the total variation (TV) regularization method, which is extensively used in image denoising. Kamilov \cite{kamilov2017parallel} further studied the basic TV minimization and proposed a novel parallel proximal-gradient method. Mallat \cite{mallat1993matching} proposed the idea of sparse representation and sparse decomposition using over-complete dictionary firstly. The K-SVD method was proposed by Elad \cite{Elad2006Image} to solved equation including learned dictionaries, which applies the sparsity of image. Mairal et al.\cite{mairal2007sparse} extended the K-SVD method and utilized it to image restoration including denoising, super resolution. Meanwhile, Liu et al. \cite{liu2013weighted} extended the basic sparse dictionary representation and proposed a dictionary learning method, which proved to be very effective in image restoration. Besides, Statck \cite{starck2004redundant} proposed a morphological component analysis method, which is a sparse decomposition algorithm and can achieve good performance. Dong \cite{Dong2011Sparsity} proposed clustering-based sparse representation (CSR) model, which is very competitive in image denoising.
\par Low rank priors of image signals are also utilized in depth image denoising \cite{lu2014depth}\cite{xue2017depth}. The most classical problem employing the low rank property \cite{kalofolias2014matrix}\cite{li2017cfmda} is the matrix completion, which can be applied in collaborative filtering or recommendation system applications. Gogna et al. \cite{gogna2014split} used the nuclear norm to solve matrix recovery and achieved good performance. Meanwhile, many researches proved that the solution of low rank can be transformed to nuclear or trace norm minimization. E.Candes \cite{candes2010matrix} proposed to apply low rank property of image in denoising problem. The nuclear norm is utilized to approximately represent the low rank of one matrix, which is proved to be a good choice to solve the denoising problem. Gu \cite{gu2014weighted} expanded the nuclear norm and proposed a weight nuclear norm minimization (WNNM) in image denoising, which can achieve great performance.
\par The self-similarity property, reflecting the intrinsic relationships of one image, has also been discussed in computer vision. The NLM that firstly employ the self-similarity property was proposed by Buades \cite{kervrann2006optimal}. Due to the great denoising effects of NLM, lots of nonlocal regularization terms have been proposed. Kostadin Dabov \cite{dabov2007image} proposed a block-matching and 3D filtering (BM3D) algorithm, which could combine the benefits of spatial filtering and transform-domain filtering to preserve the details of images accurately. Meanwhile, BM3D algorithm is universally recognized as the best denoising algorithm. In addition, the graph model is also a novel idea using the self-similarity and widely applied in image reconstruction recently. Hu \cite{hu2013depth} and Liu \cite{liu2017random} used the basic graph model according to the tree distance to solve the image denoising problem and achieved great performance. The non-local graph-based transform (NLGBT) proposed by Hu \cite{hu2013depth} is also a state-of-the-art depth image denoising algorithm. Yael \cite{kervrann2006optimal} developed a study-based graph according to the intrinsic relationships of one image.
\par It is notable that the state-of-the-art methods such as BM3D and NLGBT are patch-based denoising methods. Zhang \cite{zhang2014group} proposed the group-based sparse representation in image restoration, which indicated that group-based framework is able to significantly improve the denoising performance.

\section{Algorithm}
In this Section, the proposed GNNLG would be described in detail. We will first present the group-based graph prior, and then we will describe the nuclear norm and learning graph model. Finally, the overall regularized iteration algorithm will be provided.
\begin{figure*}
  \begin{center}

    \includegraphics[width=15cm,height=5cm]{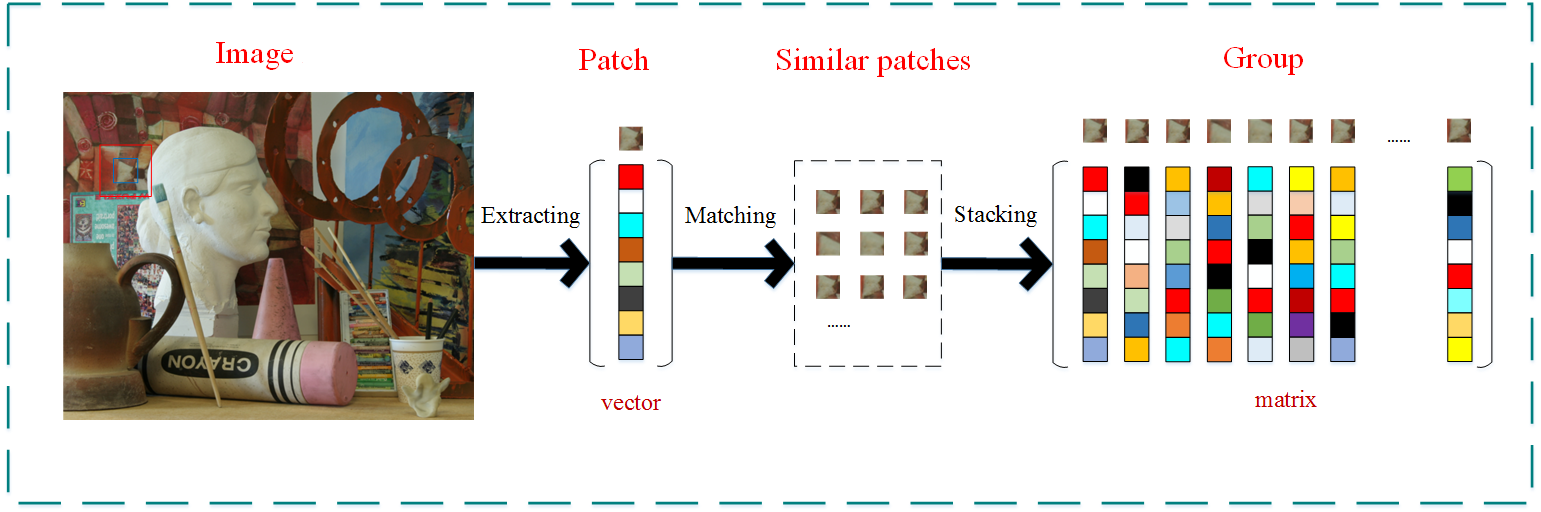}

    \caption{Illustrations for the group acquisition. Each overlapped patch $P$ (blue box) in the image $X$ is transformed to the vector format. Then the similar patches are searched in search window (red box) according to KNN algorithm. The group $T$ is constructed by stacking all the similar patches.}
  \end{center}
\end{figure*}

\subsection{Group-based Graph Prior}
\par An undirected weighted graph can be denoted by $G = (V, E, W)$, where $V$ is a vertex set composed of $N$ vertices (or nodes), $E$ is an edge set consisted of multiple $V$$\times$$V$ weighted edges, and $W$ is a weighted adjacent matrix. Generally, the weighted adjacent matrix $W$ of undirected weighted graph is non-negative and has equal diagonal elements; i,e. $W_{ij}=W_{ji}$ and $W_{ij}\geq0$. It is notable that the weighted adjacent matrix $W$ reflects the similarity between vertices $v_i$ and $v_j$. The graph can usually be represented by Laplacian matrix $L$, which plays a vital role in describing the data characteristics of a graph. In the undirected weighted graph, the computational way of $L$ is decided by $W: L = \Delta - W$, where $\Delta$ is a diagonal matrix and fulfills the equation: $\Delta_{ii} = \sum_{j}W_{ij}$. The Laplacian matrix encapsulates the graph structure for data matrix, which plays an important role in denoising. The graph term $\left \| x \right \|_{G}^2=tr(x^TLx)$ is accociated to the trace of Laplacian matrix. The formulation applying graph in denoising is as follows:
\begin{equation}
  \begin{aligned}
    \mathop{\arg\min}_{x}\frac{1}{2}{\left \| {y-x}\right \|_2^2+\theta \cdot x^TLx},
  \end{aligned}
\end{equation}
where both $x$ and $y$ are $n\times1$ vector representing the image patches, $L$ is the $n\times n$ Laplacian matrix, and $\theta$ is the regularized parameter of graph.
\par Conventional construction of graph Laplacian matrix $L$ is decided by the weighted adjacent matrix $W$. The weight matrix $W$ to the edges are commonly constructed by a threshold Gaussian kernel \cite{chaudhuri2016region}\cite{rousseau2011supervised}:
\begin{equation}
  W_{ij} =\left\{
  \begin{aligned}
    &exp(-\frac{ \left \| d_{i}-d_{j} \right \|_2^2}{\sigma^2}) \qquad if \quad \left \| d_{i}-d_{j} \right \|_2^2 \leq \epsilon \\
    &0 \qquad \qquad \qquad \qquad \quad otherwise, \\
  \end{aligned}
  \right.
\end{equation}
where $\left \| d_{i}-d_{j} \right \|_2^2$ is the Euclidean distance between two nodes $v_{i}$ and $v_{j}$, $\sigma$ is the control parameter determining how fast weights decay as the distance increases, $\epsilon$ is the threshold parameter standing for $\epsilon$-neighborhood graph.
\par Inspired by Zhang \cite{zhang2014group}, we employ the group-based graph model in denoising. We search $m$ similar patches and construct the group via block-matching for each overlapped patch of the noisy image. The group construction is described in Fig. 1. Considering that the group is a matrix, we proposed to construct the dual graph including row graph and column graph for group $T^{m \times n}$. Row graph $\left \| X \right \|_{G_r}^2 = tr(X^TL_rX)$ of the group exploits the similarity of the pixel intensities locating at the same position within all the similar patches and column graph $\left \| X \right \|_{G_c}^2 = tr(XL_cX^T)$ exploits similarity of the pixel intensities  corresponding to all the locations within each patch of the group, where $L_r^{m \times m}$ and $L_c^{n \times n}$ are the row Laplacian matrix and column Laplacian matrix, respectively. Then the expression applying the dual graph model can be formulated as follows:
\begin{equation}
  \begin{aligned}
    \mathop{\arg\min}_{X}\frac{1}{2}{\left \| {Y-X}\right \|_2^2+\theta_r \left \| X \right \|_{G_r}^2+\theta_c \left \| X \right \|_{G_c}^2},
  \end{aligned}
\end{equation}
where $X$ and $Y$ are both the $m\times n$ data matrix, $\theta_r$ and $\theta_c$ are the regularized control parameters, determing the influence degree of the regularization terms (row graph $\left \| X \right \|_{G_r}^2$ and column graph $\left \| X \right \|_{G_c}^2$), respectively.

\par To construct the dual graph, a conventional way is to regard each data matrix $X^{m \times n}$ as $n$-dimensional column vectors denoted by $X=(x_1, \cdots, x_n)$ or $m$-dimensional row vectors denoted by $X=((x'_1)^T, \cdots, (x'_m)^T)^T$. Consequently, each vector(each row or each column) in matrix is viewed as a node to compute the weighted adjacent matrix $W$. Then the Euclidean distance function of row weighted adjacent matrix $W^{m \times m}$ and column weighted adjacent matrix $W^{n \times n}$ are $\left \| (x'_{i})^T-(x'_{j})^T \right \|_2^2$ and $\left \| x_{i}-x_{j} \right \|_2^2$, respectively. Consequently, the row Laplacian matrix $L_r$ and column Laplacian matrix $L_c$ can be acquired utilizing obtained $W^{m \times m}$ and $W^{n \times n}$.
\par Conventional weighted matrix in graph Laplacian construction, as indicated in Eq. (3), considers each pixel in the data matrix as a node, which is very efficient for an arbitrary input image. As indicated in Eq. (3), conventional weighted matrix in graph Laplacian construction considers each pixel in the data matrix as a node, which explores the similarity among pixes in a patch. For reflecting the network connectivity and intrinsic relationships of similar patches well, the newly Laplacian matrix is provided in our paper. To exploit the intrinsic structure and underlying topology further, we proposed a learning based graph Laplacian construction method. In this paper, the acquisition of the learned Laplacian matrix is derived from an optimized formulation related to the graph. To construct the graph, the regularization term over $L$ should be added in our formulation besides the graph $\left \| X \right \|_{G}^2$:
\begin{equation}
  \begin{aligned}
    & \mathop{\arg\min}_{L} \alpha \left \| X \right \|_{G}^2+\beta \left \| L \right \|_F^2 \\
    & s.t. ~~ L_{ij}=L_{ji}\leq 0 \\
    & \qquad L\cdot1=0 \\
    & \qquad Tr(L) = N, \\
  \end{aligned}
\end{equation}
where $\alpha$, $\beta$ are the control parameters and $N$ is the number of graph nodes.
It is notable that the Laplacian matrix $L$ is vectorized in our objective function, which denotes that Eq. (5) is a quadratic optimization problem. Meanwhile, the optimization problem includes three constraints. The first constraint ensures that the Laplacian matrix is valid and the second one makes sure that the result is more effective. The third constraint describes that the trace is equal to the node of $L$ according to the mathematical meaning. For each data matrix $X^{m \times n}$, the number of node $N$ in the optimization problem is equal to $m$ or $n$ when learning the row Laplacian matrix $L_r$ or column Laplacian matrix $L_c$ \cite{Yankelevsky2017Dual}.
\par Overall, Eq. (5) employs graph learning by optimizing the graph model and the Laplacian matrix. Meanwhile, it is notable that the learned graph Laplacian is self-adaptive and ensures high smoothness for each data matrix.

\subsection{Nuclear Norm and Learning Graph Model}
Considering that the group $T$ has strong low rank property, we further introduce the low rank prior in our work. Generally, a regular replacement to low rank of data matrix $X$ is known as nuclear norm or trace norm $\left \|X\right \|_* = tr((XX^T)^{1/2})=\Sigma_k{\sigma_k}$, where $\sigma_k$ are the singular values of $X$. Consequently, the combination of group-based nuclear norm and graph model can be expressed as follows:
\begin{flalign}
  \min_{X}\theta_{n} \left \| X \right \|_*+\frac{1}{2}\left \| {X-T}\right \|_2^2+\theta_{r}\left \| X \right \|_{G_r}^2+\theta_{c}\left \| X \right \|_{G_c}^2,
\end{flalign}
where  $\theta_n$, $\theta_r$ and $\theta_c$ are the control parameters of nuclear norm, row graph and column graph respectively. The graph regularization term reflects the nonlocal self-similarity and the nuclear reflects the low rank property of image, which can utilize much information of image.
\par Considering both the accuracy and complexity, we employ the alternating direction method of multipliers (ADMM) to solve Eq. (6). The ADMM is a convex optimization algorithm that could blend the decomposability of dual ascent and the superior convergence properties of multipliers \cite{boyd2011distributed}. Using ADMM, Eq. (6) can be represented as:
\begin{equation}
  \begin{aligned}
    & \min_{X}\theta_{n} \left \| X \right \|_*+\frac{1}{2}\left \| {Z-T}\right \|_2^2+\theta_{r}\left \| Z \right \|_{G_r}^2+\theta_{c}\left \| Z \right \|_{G_c}^2
    \\
    & s.t. \quad X=Z.
  \end{aligned}
\end{equation}

\par In order to process Eq. (7) better, we let $F(X)=\theta_{n} \left \| X \right \|_*$, $G(Z) = \frac{1}{2}\left \| {Z-T}\right \|_2^2+\theta_{r}\left \| Z \right \|_{G_r}^2+\theta_{c}\left \| Z \right \|_{G_c}^2$, then the augmented Lagrangian of our method can be expressed as:
\begin{equation}
  \begin{aligned}
    L_p(X,Z,Y)= & F(X)+G(Z)+ Y^T(X-Z) \\
    & +\frac{p}{2}\left \| X-Z \right \|_2^2,
  \end{aligned}
\end{equation}
where $Y$ is the dual variable and $p$ is the augmented Lagrangian parameter.
\par Using $k$ as the iteration number, the relevant iterations composed of $X$-subproblem, $Y$-subproblem, and $Z$-subproblem steps can be acquired as:
\begin{equation}
  \begin{aligned}
    & X^{k+1}= \mathop{\arg\min}_{X}L_p(X, Z^k, Y^k), \\
    & Z^{k+1}= \mathop{\arg\min}_{Z}L_p(X^{k+1}, Z, Y^k), \\
    & Y^{k+1}=Y^k+X^{k+1}-Z^{k+1}.
  \end{aligned}
\end{equation}
\par Eventually, Eq. (6) is transformed into the solution of three sub-optimization problems as can be seen in the Eq. (9). The roles of $X$ and $Z$ are almost symmetric and dual updated. To solve Eq. (9) more intuitively, three proposed subproblems can be described using Eq. (10)-(12):
\begin{flalign}
  & X^{k+1}= \mathop{\arg\min}_{X}(F(X)+\frac{p}{2}\left \| X-Z^k+Y^k \right \|_2^2), \\
  & Z^{k+1}= \mathop{\arg\min}_{Z}(G(Z)+\frac{p}{2}\left \| X^{k+1}-Z+Y^k \right \|_2^2), \\
  & Y^{k+1}=Y^k+X^{k+1}-Z^{k+1}.
\end{flalign}
\par Then the combination problem is optimized by three subproblems. The detailed solution to $X$-subproblem, $Z$-subproblem and stopping criterion of our ADMM algorithm are as follows:
\subsubsection{$X$-Subproblem}
\par The expression in Eq. (10) is actually a nuclear norm with a data term. It has a proximate form and can be transformed to the singular value decomposition (SVD) of $Z-Y$. Then the proximate solution to $X$-minimization  is:
\begin{equation}
  \begin{aligned}
    X^{k+1}&= \mathop{\arg\min}_{X}\theta_{n} \left \| X \right \|_*+\frac{p}{2}\left \| X-Z^k+Y^k \right \|_2^2 \\
    &= U \cdot \Gamma_{\lambda,v}(\Sigma(Z^k-Y^k)) \cdot V ^T,
  \end{aligned}
\end{equation}
where $Z^k-Y^k=U\cdot \Sigma \cdot V^T$ is the SVD of $Z^k-Y^k$. It is notable that $\Gamma_{\lambda,v}$ is the fast threshold algorithm, where $\lambda$ and $v$ are two threshold parameters. In Eq. (13), $\lambda$ is equivalent to $\theta_n / p$. The operator $\Gamma_{\lambda,v}$ is defined as follows:
\begin{equation}
  \begin{aligned}
    \Gamma_{\lambda,v}(x)=max(0, \left| x \right| - \lambda\left| x \right|^{v-1})\frac{x}{\left| x \right|}.
  \end{aligned}
\end{equation}
\par It is noteworthy that the proposed fast threshold algorithm amounts to conventional soft threshold when $v$  is close to 1.  Meanwhile, when $v$ approaches to 0, Eq. (14) is approximated to hard threshold algorithm \cite{kamilov2017parallel}. The vivid description of the proposed threshold algorithm can be seen in Fig. 2.

\begin{figure}[h]
  \begin{center}

    \includegraphics[width=9cm,height=6cm]{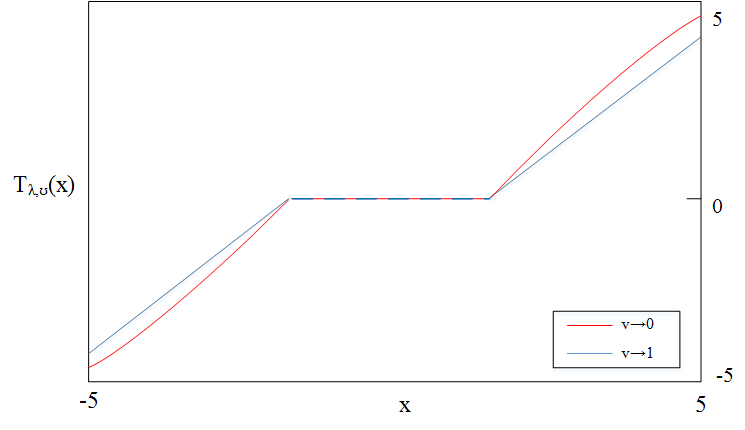}

    \caption{The vivid description of the latest fast threshold algorithm $\Gamma_{\lambda, v}$. When $v$ is closed to 1, $\Gamma_{\lambda, v}$ amounts to the conventional soft threshold method. When $v$ approaches to 0, $\Gamma_{\lambda, v}$ is similar to the hard thresholding.}
  \end{center}
\end{figure}

\subsubsection{$Z$-Subproblem}
\par Unlike $X$-minimization, the $Z$-minimization doesn't have a close convex form to solute. But we can compute the derivative to make $Z$ minimize. After transformation, the $Z$-minimization can be solved by the following formulation:
\begin{equation}
  \begin{aligned}
    2\theta_r L_r\otimes Z&+2\theta_c Z\otimes L_c+(1+p)Z \\
    &=T+p(X^{k+1}+Y^k).
  \end{aligned}
\end{equation}
where $\otimes$ is the Kronecker product. Actually, $X^{k+1}$ is close to $Z^{k+1}$ infinitely when the iteration number is big enough \cite{kalofolias2014matrix}.

\subsubsection{The stopping criterion of ADMM}
The stopping criterion is derived from the convergence of ADMM algorithm. In our solution, the stopping criterion needs to fullfill the dual and primal feasibility condition. And we ensure that the small iterative dual residual $s^{k+1}$ and primal residual $r^{k+1}$ satisfy the feasibility condition. The specific stopping criterion is as follows:
\begin{equation}
  \begin{aligned}
    &\left \| s^{k+1} \right \|=p(Z^{k+1}-Z^{k})\leq \epsilon^{dual} \\
    &\left \| r^{k+1} \right \|=X^{k+1}+Z^{k+1}\leq \epsilon^{pri},
  \end{aligned}
\end{equation}
where $\epsilon^{dual}$ and $\epsilon^{pri}$ are the feasibity tolerances for the dual and primal feasibility conditions, respectively.

\subsection{Overall Regularized Iteration}
\par From Fig. 1, it is noteworthy that the minimum unit of our algorithm is the overlapped patch $P$. The main processing object of our algorithm is the group $T$ composed of similar patches. After each overlapped patch is processed by the proposed method, we use the weighted averaging method to update our denoising result.
\par Inspired by the works in  \cite{hu2013depth}\cite{hao2014iterative}, we proposed an iterative regularization to improve the quality of denoised image further. The homologous formulation is as follows:
\begin{equation}
  \begin{aligned}
    X_{k+1}=X_k+\delta(X-X_k),
  \end{aligned}
\end{equation}
where $X$ is the original noisy image, $X^k$ is the denoised result after the $k$-th iteration and $\delta$ is the regularized parameter.
\par
Overall, our proposed GNNLG can be solved according to above steps. The summary of our GNNLG algorithm is provided in \textbf{Algorithm 1}.

\begin{algorithm}
  \caption{Depth image denoising using GNNLG}
  \KwIn{original depth image $X$, noisy depth image $X_{noisy}$}\ 
  \KwOut{final denoised depth image $X'_{final}$}\ 
  \textbf{Initialization}: $X'_{0}$ = $X_{noisy}$, $X_{0}$ = $X_{noisy}$\; 
  \For{$i \leftarrow 1$ \KwTo outer iteration number $n_1$}{
    Step1: Perform iterative regularization by Eq. (17) \\
                 $  $ \quad \quad \, to acquire $X_{i}$\;
    Step2: Divide $X_{i}$ in $k$ overlapped patches\;
    Step3: Acquire revelant group $T$ for each patch\;
    Step4: Learn the graph Laplacian matrix for each \\
                 $  $ \quad \quad \, patch by Eq. (5)\;
    Step5: Solve the Combination Eq. (6) for each patch \\
                $  $ \quad \quad \, by ADMM:\\
    \While{$j$ $\le$ inner iteration number $n_2$}{

      \eIf{the stopping criteria is not reached}{
        Solve X subproblem by Eq. (13)\;
        Solve Z subproblem by Eq. (15)\;
        Solve Y subproblem by Eq. (12);
      }{
      break;
    }
    $j \leftarrow j+1$
  }
  Step6: Update image $X'_{i}$ for $k$ overlapped denoised \\
            $  $ \quad \quad \, patches by weighted averaging method\;
  $X_{i} \leftarrow X'_{i}$ \\
  $i \leftarrow i+1$\\
}
$X'_{final}$ = $X'_{i}$ when PSNR($X$, $X'_{i}$) is the maximun value.
\end{algorithm}

\section{Results}
In this section, the experimental results would be presented and discussed to verify that our proposed method is superior to other state-of-the-art methods. The depth images we use are $Art$ (463$\times$370), $Books$ (463$\times$370), $Dolls$ (695$\times$555), $Moebius$ (463$\times$370), $Reindeer$ (447$\times$370), $Laundry$ (653$\times$555), $Cones$ (350$\times$375) and $Teddy$ (450$\times$375). The noisy image is derived by adding the additive white Gaussian noise (AWGN) to the original clean depth images. The level of added noise is measured by the standard deviation $\sigma$ ranging from 15 to 30.The peak signal-to-noise ratio (PSNR) \cite{Gu2015Using} is adopted to evaluate the quality of denoised image in our experiments. It is notable that all of our experiments are operated in Matlab 2014b on a Dell computer with Intel(R) Ceon(R) CPU E5-2623 Windows processor (3GHz), 64 G random access memory and Winodws 10 operating system.Because of the long processing time of nuclear norm, our computational time and complexity is similar with weight nuclear norm minimization \cite{gu2014weighted}.

\subsection {Parameters Setting}
\par The parameters in our model mainly exist in block matching, graph construction and iterative scheme. The concrete setting of parameters are as follows: in the stage of block matching, the basic patch size is 5$\times$5. The search window is set to 20$\times$20 and the search intervals between patches are set to 3 pixels. The number $K$ of similar patches is 16. In the stage of graph Laplacian learning, the control parameters $\alpha$ and $\mu$ are set to 1.2 and 0.8 respectively. In the stage of combining nuclear norm and graph model, the parameters $\theta_n$, $\theta_r$ and $\theta_c$ are not fixed. The final denoising effects are depended on these three parameters. The parameter $v$ in fast threshold algorithm is set to 0.1 and the parameter $p$ in ADMM algorithm is set to 0.015. In the stage of overall iterative regularization, the regularized parameter $\delta$ is set to 0.1. The outer iterative number $n_1$ and inner iterative number $n_2$ are set to 5 and 100, respectively. The dual feasibility $\epsilon^{dual}$ and primal $\epsilon^{pri}$ feasibility tolerances conditions are default in ADMM algorithm and are set to 0.015, 0.03, respectively.
\par In fact, the most important parameter are the regularization terms in our experiment. Meanwhile, we set these three parameters by controlling variable method. When setting one parameter such as $\theta_n$, we set other two parameters $\theta_r$ and $\theta_c$ are 0. For depth image $Art$, these three concrete parameters setting are provided in the Fig. 3, and we can find these parameter value are gradually increased with the noise variable rised.

 \begin{figure*}
 \begin{center}
 \includegraphics[width=4.5cm,height=4.2cm]{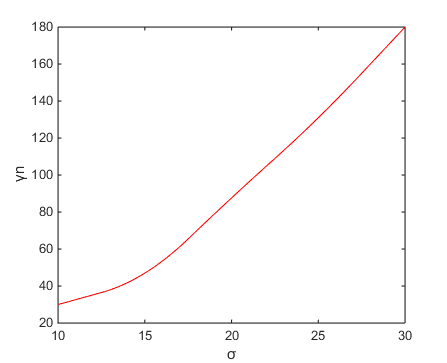}
 \includegraphics[width=4.5cm,height=4.2cm]{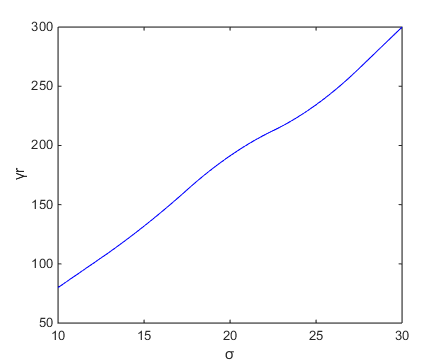}
 \includegraphics[width=4.5cm,height=4.2cm]{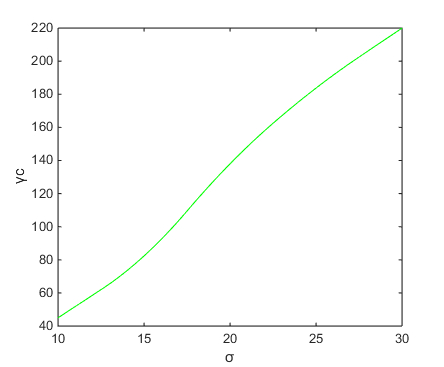}
 \caption{ The comparison of regularization terms value change curves under different AWGN over depth images $Art$.}
 \end{center}
 \end{figure*}

\begin{table*}
\caption{The PSNR (dB) and SSIM comparisons over eight depth images by seven representative methods with the AWGN variable $\sigma$ = 15 and $\sigma$ = 20.}
\renewcommand\arraystretch{1.1}
\centering
\resizebox{15cm}{4cm}{
\begin{tabular}{|m{0.3cm}<{\centering}|m{1.2cm}<{\centering}|m{2cm}<{\centering}|m{2cm}<{\centering}|m{2cm}<{\centering}|m{2.3cm}<{\centering}|m{2.2cm}<{\centering}|m{2.3cm}<{\centering}|}
\Xhline{1.2pt}

\multirow{2}{*}{$\sigma$}&\multirow{2}{*}{Image}&\multicolumn{6}{c|}{Methods}\\
\cline{3-8}
& &  NLM & CSR & NLGBT &BM3D & WNNM & GNNLG\\

\hline    \multirow{8}{*}{15}& Art& 38.98 / 0.9608&37.75 / 0.9621&38.76 / 0.9598&38.43 / 0.9769&38.98 / 0.9762&\textbf{39.53 / 0.9782}\\
\cline{2-8}      &Books   & 39.34 / 0.9769&40.60 / 0.9705 & 40.85 / 0.9659& \textbf{42.57 / 0.9787}& 41.93 / 0.9797&    41.96 / 0.9763\\
\cline{2-8}        &Cones  & 36.96 / 0.9668 & 37.44 / 0.9633 & 39.24 / 0.9706 & 38.57 / 0.9708 & 39.36 / 0.9570 &  \textbf{39.64 / 0.9783}\\
\cline{2-8}       &Dolls & 36.84 / 0.9662 &37.98 / 0.9645   & 38.08 / 0.9658 & 39.59 / 0.9769 & 40.19 / 0.9670 & \textbf{40.53 / 0.9688}\\
\cline{2-8}        &Laundry  & 39.54 / 0.9533 &39.46 / 0.9582 & 39.21 / 0.9562 & 41.62 / 0.9734 & 40.08 / 0.9662 &\textbf{41.97 / 0.9683}\\
\cline{2-8}        &Moebius & 38.69 / 0.9692&39.16 / 0.9628  & 39.71 / 0.9688 & 40.71 / 0.9731& 40.28 / 0.9797&\textbf{41.37 / 0.9776}\\
\cline{2-8}        &Teddy  & 37.58 / 0.9721&35.86 / 0.9683   & 39.17 / 0.9647 & 40.44 / 0.9758 & 39.18 / 0.9765 &\textbf{40.53 / 0.9711}\\
\cline{2-8}        &Reindeer & 40.67 / 0.9687& 39.55 / 0.9632 & 40.34 / 0.9692 & 41.26 / 0.9714 & 40.64 / 0.9727 &\textbf{41.54 / 0.9736}\\
\cline{1-8}        &\textbf{Average} &  38.58 / 0.9667 &  38.48 / 0.9641 &  39.67 / 0.9651  &40.31 / 0.9736 & 40.08 / 0.9719 &  \textbf{40.88 / 0.9740}
\\

\hline    \multirow{8}{*}{20}& Art & 36.93 / 0.9533 & 35.84 / 0.9501   & 36.00 / 0.9488 &36.80 / 0.9548 & 36.93 / 0.9531 &\textbf{37.38 / 0.9620}    \\
\cline{2-8}      &Books  & 38.29 / 0.9703 &38.35 / 0.9665 & 38.77 / 0.9690 & 39.58 / \textbf{0.9723}& 39.54 / 0.9711 &\textbf{39.98} / 0.9721     \\
\cline{2-8}        &Cones & 35.83 / 0.9530 &35.54 / 0.9483& 36.45 / 0.9488 & 36.21 / 0.9540 & 37.03 / 0.9574&\textbf{37.56 / 0.9577}  \\
\cline{2-8}       &Dolls & 35.91 / 0.9511&36.66 / 0.9528   & 37.66 / 0.9616 & 36.57 / 0.9602 & 38.06 / 0.9632 &\textbf{38.51 / 0.9655} \\
\cline{2-8}        &Laundry & 38.47 / 0.9668 &38.27 / 0.9622 & 37.29 / 0.9595 & 39.88 / 0.9680 & 38.33 / 0.9662& \textbf{40.10 / 0.9713}\\
\cline{2-8}        &Moebius  & 37.51 0.9646 &37.11 / 0.9590 & 37.52 / 0.9547 & 38.69 / 0.9612 & 38.58 / 0.9622 &\textbf{39.15 / 0.9684}\\
\cline{2-8}        &Teddy   & 36.67 / 0.9593 & 35.86 / 0.9557  & 36.98 / 0.9403 & 37.70 / 0.9548 & 37.27 / \textbf{0.9548} &\textbf{38.00} / 0.9517\\
\cline{2-8}        &Reindeer   & 39.20 / 0.9710 &37.35 / 0.9669   & 37.74 / 0.9633 & 37.19 / 0.9684 & 38.62 / 0.9692 &\textbf{39.93 / 0.9726}\\
\cline{1-8}        &\textbf{Average } &37.35 / 0.9611   &36.87 / 0.9577 & 37.30 / 0.9557 &37.83 / 0.9616 &  38.05 / 0.9621& \textbf{38.83 / 0.9652} \\

\Xhline{1.2pt}
\end{tabular}
}
\end{table*}

\begin{figure}[!htpb]
\center
\includegraphics[height=3.72in, angle=0]{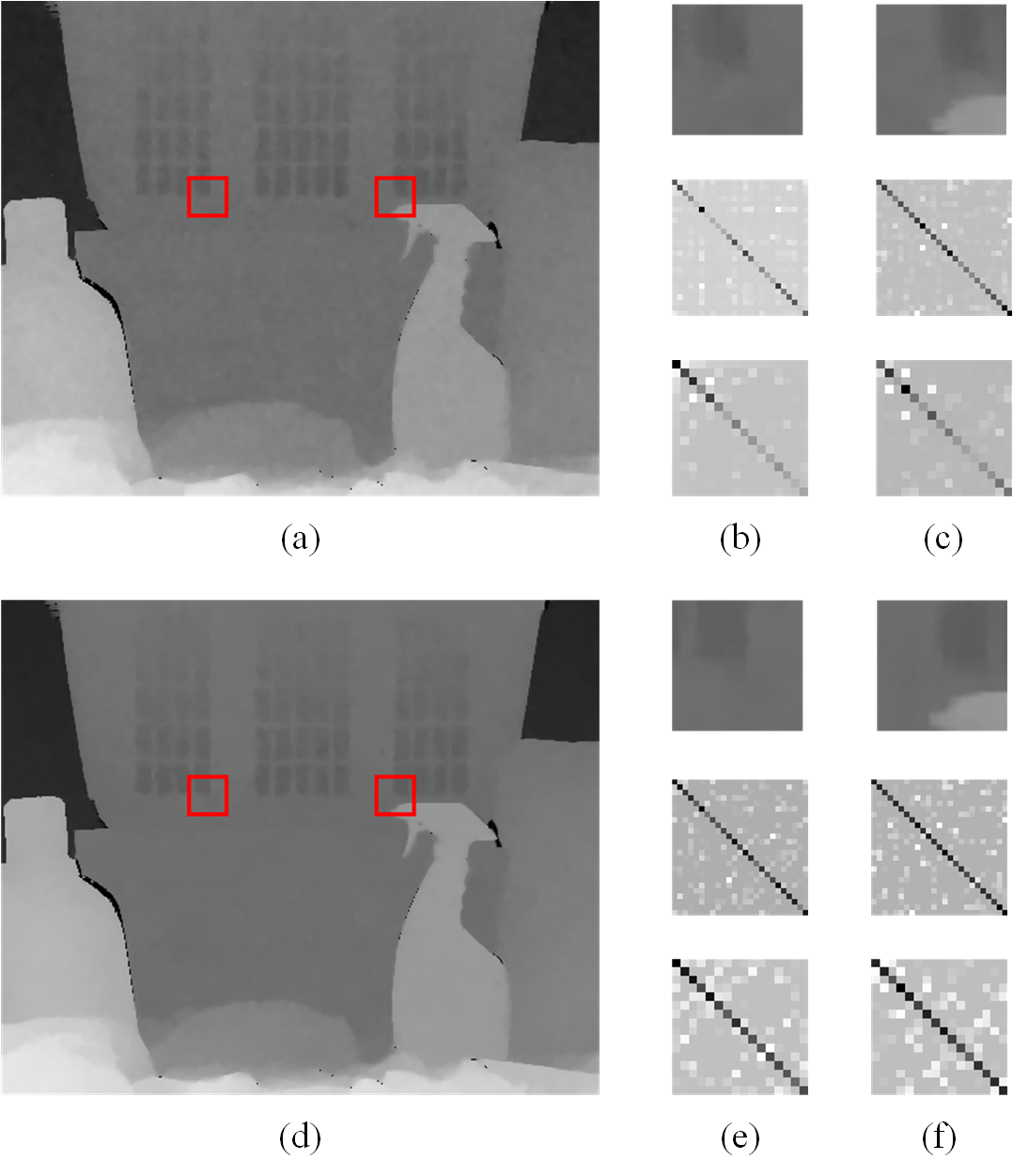}
\caption{Visualization of the denoising results and corresponding row Laplacian matrix as well as column Laplacian matrix comparison for depth image $Laundry$. (a) and (d) are the denoising results using the conventional graph and proposed learning graph model, respectively; (b) and (e) are the magnified denosiing result (top), row Laplacian matrix (middle) and column Laplacian matrix (bottom) corresponding to the left red box in (a) and (d), respectively; (c) and (f) are the magnified denosing result (top), row Laplacian matrix (middle) and column Laplacian matrix (bottom) corresponding to the right red box in (a) and (d), respectively. }
\end{figure}

\subsection {Analysis of Graph Laplacian Matrix}
\par To reflect the superior performance of the learned graph, the denoising results and corresponding row Laplacian matrix as well as column Laplacian matrix comparison between the conventional method and the proposed learning graph model over depth image $Laundry$ are provided in Fig. 4, where the noise variable $\sigma$ is 20. Here, we normalize the magnitudes of the Laplacian matrix firstly. Then we acquire the magnitude of the Laplacian matrix and we represent it as a gray image. The black region of the Laplacian matrix represents the larger magnitude, and the white region represents the smaller magnitude.

As shown in Fig. 4 (a) and Fig. 4 (d), the denoising result by learned graph Laplacian matrix removes almost all the noises and preserves more details than the conventional graph Laplacian matrix. To further exhibit the difference between Fig. 4 (a) and Fig. 4 (d), we select two sample boxes as indicated by the red rectangule and give the magnified visualization in the top row of Fig. 4 (b), Fig. 4 (c), Fig. 4 (e), and Fig. 4 (f), respectively. We can clearly observe that the results generated by the learned graph Laplacian matrix maintain the sharper edge and exhibit better visual quality. Besides, for the regions indicated by the right rectangule, the blurring effect in the top row of Fig. 4 (c) is thoroughly removed in the top row of Fig. 4 (f), which indicates the superiority of the proposed learning graph model.

Another observation is that the regions along the diagonal direction in the learned graph Laplacian matrix (middle and bottom images of Fig. 4 (e) and Fig. 4 (f)) are much darker and have more distinct textures than those in the conventional graph Laplacian matrix (middle and bottom images of Fig. 4 (b) and Fig. 4 (c)), which indicates that the magnitudes of diagonal in the learned graph Laplacian matrix are much larger than those in the conventional Laplacian matrix. Such a property is very benefical to maintain the sharpness for the edge regions. This is consistent with the fact that the blurring effects around the edge regions in the top row of Fig. 4 (c) can be greatly removed in the top row of Fig. 4 (f).

\subsection{Convergence Analysis}
In this subsection, the convergence of the overall optimized iteration is analyzed for the proposed GNNLG algorithm.  The regularized iterative results of depth image $Art$ and $Cones$ are provided in Fig. 5. We can see that the PSNR would get improved with the increase of iteration number, which shows the superiority of overall iteration. More specifically, when $\sigma$ = 10, our overall iteration is most effective, which can greatly increase the PSNR of depth image. Furthermore, no matter the noise variable $\sigma$ is 20 or 30, the PSNR of depth image is improved after two or three iterations and retain a stable value when the iteration number increases for both depth image $Art$ and $Cones$. These indicate that the overall iteration method can improve the image quality further and achieve the great convergence in our work.

\subsection{Performance Evaluation}
\par To prove the superiority of our proposed GNNLG, we compare our method with several state-of-the-art denoising methods: Total Variation \cite{selesnick2017total}, Non-Local Means \cite{kervrann2006optimal}, Block-Matching and 3D filtering \cite{dabov2007image}, Weighted Nuclear Norm Minimization \cite{gu2014weighted}, Non-local Graph-Based Transform \cite{hu2013depth}, Clustering-based Sparse Representation \cite{Dong2011Sparsity}. Both objective and subjective visual quality evaluation are provided in our experiment.
\par The concrete objective quality comparisons among different denoising methods are shown in Table \uppercase\expandafter{\romannumeral1} and Table \uppercase\expandafter{\romannumeral2}. As can be seen in the Table \uppercase\expandafter{\romannumeral1}, when $\sigma$ = 15, the PSNR of seven test depth images except the image $book$ by using our GNNLG model is higher than other methods. More specifically, the average PSNR gains of our GNNLG compred to the second best one, the BM3D, can be up to 0.57dB.
When $\sigma$ = 20, the PSNR results of GNNLG over all the test images are the highest among the seven competing methods. Meanwhile, the PSNR gains of GNNLG is up to 0.6dB compared with the second best method, WNNM, in average. Furthermore in Table  \uppercase\expandafter{\romannumeral2}, the PSNR results produced by our GNNLG model are higher than all the other methods over the eight depth images for both $\sigma$ = 25 and $\sigma$ = 30. The PSNR gains of our GNNLG compared with the scond best one are 0.78dB and 0.68dB for $\sigma$ = 25 and $\sigma$ = 30, respectively.
In a word, more noise added in the image, more effective our GNNLG will be, which revals the GNNLG is much robust and competitive compared with other seven representative denoising methods

\par Besides the objective evaluation over the eight depth images, the subjective quality of denoising results over four depth images including $Moebius$, $Teddy$, $Art$ and $Cones$  generated by six methods are provided in Fig. 6 to Fig. 9. All these figures provide the intuitionistic visual denoising results of all the images with different AWGN varying from 15 to 30. In Fig. 6, the NLM method exhibits the obvious blurry result, and the CSR method gets the same blurring result like NLM method, NLGBT produces the contour jaggies, which destroys the details of image. BM3D and WNNM achieve better perofrmance than other compared methods. GNNLG produces more smooth and cleaner denoising result than other methods. Meanwhile, our GNNLG preserves more detailed information and distinct image textures than BM3D and WNNM for depth image $Moebius$. In Fig. 7, except the NLGBT, WNNM and GNNLG, other method can't obtain the edge details greatly; but NLGBT produces the dirty pixels and the WNNM produces the unnatural image. In contrast, our GNNLG gets more smooth result than WNNM and NLGBT. In Fig. 8 and Fig. 9, almost all the methods except NLGBT and GNNLG produce the blurry and dirty results for image $Art$ and $Cones$. It is notable that NLGBT did not remove the noise thoroughly. However, our GNNLG produces the great denoising result, which both preserves the important details and achieves much sharper edges. These visual effects also validate the superiority of our proposed method exploiting the combination of low rank and nonlocal similarity.

\par Overall, our GNNLG model is superior to other state-of-the-art methods no matter in subjective or objective quality criterion from these Figures and Tables.


%

\begin{table*}
\caption{The PSNR (dB) and SSIM comparisons over eight depth images by seven representative methods with the AWGN variable $\sigma$ = 25 and $\sigma$ = 30.}
\renewcommand\arraystretch{1.1}
\centering
\resizebox{15cm}{4cm}{
\begin{tabular}{|m{0.3cm}<{\centering}|m{1.2cm}<{\centering}|m{2cm}<{\centering}|m{2cm}<{\centering}|m{2cm}<{\centering}|m{2.2cm}<{\centering}|m{2cm}<{\centering}|m{2.3cm}<{\centering}|}
\Xhline{1.2pt}
\multirow{2}{*}{$\sigma$}&\multirow{2}{*}{Image}&\multicolumn{6}{c|}{Methods}\\
\cline{3-8}
& &  NLM & CSR & NLGBT &BM3D & WNNM & GNNLG\\

\hline    \multirow{7}{*}{25}& Art & 35.23 / 0.9478 &34.23 / 0.9481   & 34.34 / 0.9460 &35.83 / \textbf{0.9531} & 35.29 / 0.9496 &\textbf{36.32} / 0.9523    \\
\cline{2-8}      &Books & 37.00 / 0.9632&36.74 / 0.9549 & 36.83 / 0.9511 & 37.83 / 0.9566& 38.13 / 0.9643&\textbf{38.54 / 0.9626}    \\
\cline{2-8}        &Cones  & 34.65 / 0.9496&33.79 / 0.9490 & 34.58 / 0.9436 & 34.63 / 0.9493 & 35.20 / 0.9476 &\textbf{35.71 / 0.9511}  \\
\cline{2-8}       &Dolls   & 34.96 / 0.9511&34.93 / 0.9428   & 35.93 / 0.9447& 35.22 / 0.9541 & 36.30 / 0.9498 &\textbf{36.86 / 0.9528} \\
\cline{2-8}        &Laundry & 37.41 / 0.9516&36.27 / 0.9483 & 35.94 / 0.9471& 38.32 / 0.9492&36.77 / 0.9467&\textbf{38.93 / 0.9503} \\
\cline{2-8}        &Moebius  &36.42 / 0.9432 &35.98 / 0.9529& 35.63 / 0.9468& 37.25 / 0.9543& 36.70 / 0.9537&\textbf{38.06 / 0.9533}\\
\cline{2-8}        &Teddy & 35.56 / 0.9473&34.48 / 0.9488  & 35.15 / 0.9413& 35.80 / \textbf{0.9528} & 35.22 / 0.9526&\textbf{36.22} / 0.9494\\
\cline{2-8}        &Reindeer &37.75 / 0.9503 &35.98 / 0.9519   & 36.08 / 0.9483 & 37.78 / 0.9543 & 36.95 / 0.9511 &\textbf{38.20 / 0.9594}\\
\cline{1-8}        &\textbf{Average}&   36.12 / 0.9505  &35.51 / 0.9458 &   35.56 / 0.9449 &  36.58 / 0.9517 &  36.32 / 0.9506 &  \textbf{37.36 / 0.9539}  \\

\hline      \multirow{8}{*}{30}& Art  & 33.89 / 0.9463 &33.12 / 0.9413    & 32.55 / 0.9316 &35.21 / \textbf{0.9477} &34.21 / 0.9426 &\textbf{35.54} / 0.9436    \\
\cline{2-8}      &Books  & 35.84 / 0.9524 &35.54 / 0.9447   & 35.06 / 0.9467 & 36.76 / 0.9463 & 36.81 / 0.9493 &\textbf{37.42 / 0.9525}    \\
\cline{2-8}        &Cones  & 33.57 / 0.9384 &32.55 / 0.9377   & 32.84 / 0.9322 & 33.66 / 0.9428 & 33.92 / 0.9415 &\textbf{34.72 / 0.9484}  \\
\cline{2-8}       &Dolls  & 34.14 / 0.9407 &33.69 / 0.9351   & 34.35 / 0.9449 & 34.58 / 0.9403 & 34.80 / 0.9482 &\textbf{35.22 / 0.9412} \\
\cline{2-8}        &Laundry   & 36.45 / 0.9543 &35.23 / 0.9463 & 34.63 / 0.9496 & 37.54 / \textbf{0.9598} &35.90 / 0.9420 &\textbf{37.87} / 0.9506 \\
\cline{2-8}        &Moebius &35.46 / 0.9432 & 34.54 / 0.9474  & 34.41 / 0.9487 & 36.22 / 0.9433 & 35.47 / 0.9462&\textbf{36.46 / 0.9486}\\
\cline{2-8}        &Teddy  & 34.45 / 0.9488 &33.35 / 0.9441   & 33.55 / 0.9378 & 34.45 / 0.9432 & 34.00 / 0.9401&\textbf{35.13 / 0.9428}\\
\cline{2-8}        &Reindeer &36.43 / 0.9469 &34.76 / 0.9401 & 34.37 / 0.9498 &35.73 / 0.9437 & 36.73 / 0.9489 & \textbf{37.23 / 0.9531} \\
\cline{1-8}        &\textbf{Average}  &35.03 / 0.9426 & 34.10 / 0.9458 &  33.97 / 0.9439 &  35.52 / 0.9446& 35.23 / 0.9423& \textbf{36.20 / 0.9463} \\

\Xhline{1.2pt}
\end{tabular}
}
\end{table*}

\begin{figure*}
\begin{center}
\includegraphics[width=6.5cm,height=5.5cm]{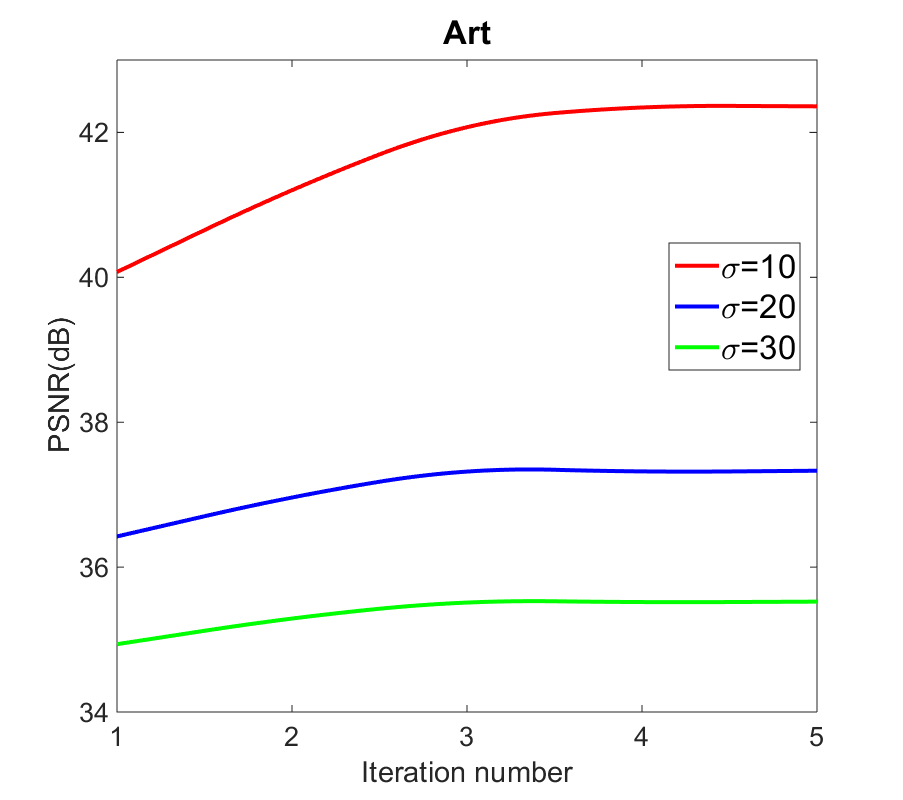}
\hspace{1cm}
\includegraphics[width=6.5cm,height=5.5cm]{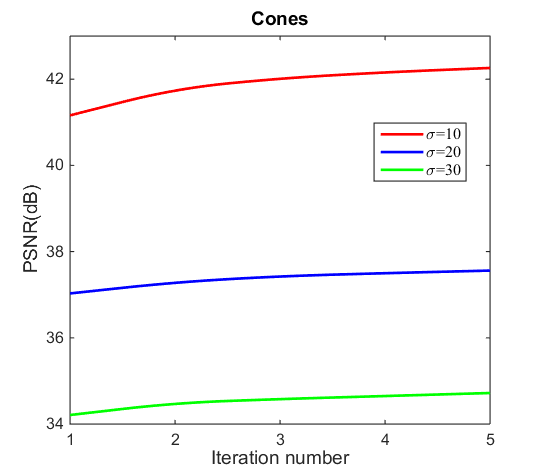}
\caption{ The comparison of PSNR curves under different AWGN over depth images $Art$ and $Cones$.}
\end{center}
\end{figure*}

\section{Conclusion}
This paper builds a novel framework for depth image denoising using the group-based nuclear norm and learning graph(GNNLG) model, which exploits the intrinsic low-rank and self-similarity property of depth image. The graph learning technique based regularization is used to construct the Laplacian matrix, which is more effective than the general construction based on tree distance in describing the intrinsic self-similarity. To achieve fast speed and high convergence, the alternating direction method of multipliers is employed to solve our non-convex combination formulation. The experimental results exhibited the denoising effects and proved that our proposed model can achieve best PSNR compared to many current state-of-the-art denoising techniques. Meanwhile, the high convergence of our model was provided, which proved that the designed algorithm is stable and self-adaptive.
\par In the future, the dictionary learning technique would be added to extend our model. Meanwhile, the deep learning method would be adopted to solve the subproblem for faster speed. Furthermore, we would explore many other applications for our proposed GNNLG, such as image super resolution and image deblurring.

\begin{figure*}
\centering
\subfigure[Src image / PSNR]{\includegraphics[width=1.4in, angle=0]{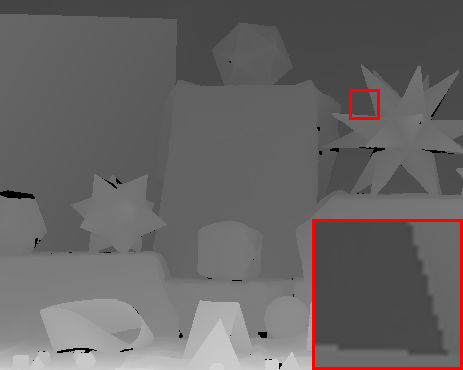}}
\subfigure[Noisy image / 24.60dB]{\includegraphics[width=1.4in, angle=0]{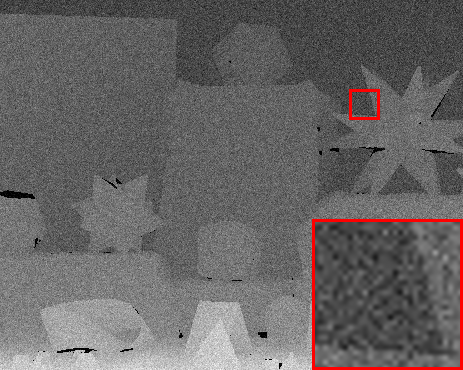}}
\subfigure[NLM / 38.69dB]{\includegraphics[width=1.4in, angle=0]{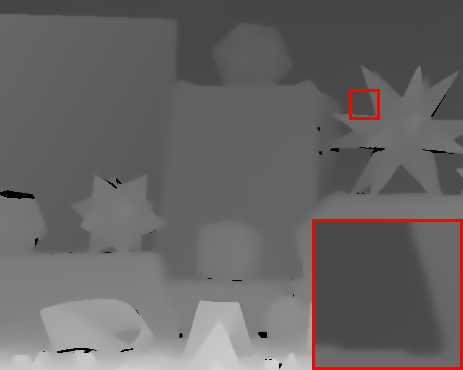}}
\subfigure[CSR / 39.16dB]{\includegraphics[width=1.4in, angle=0]{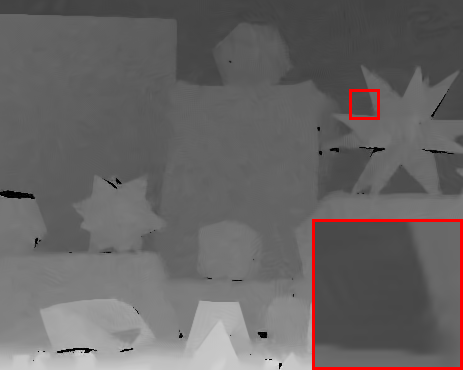}}
\subfigure[NLGBT / 39.71dB]{\includegraphics[width=1.4in, angle=0]{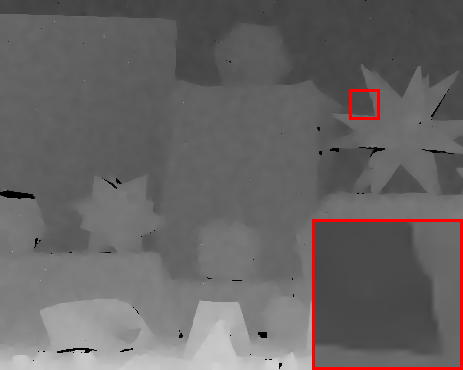}}
\subfigure[BM3D / 40.71dB)]{\includegraphics[width=1.4in, angle=0]{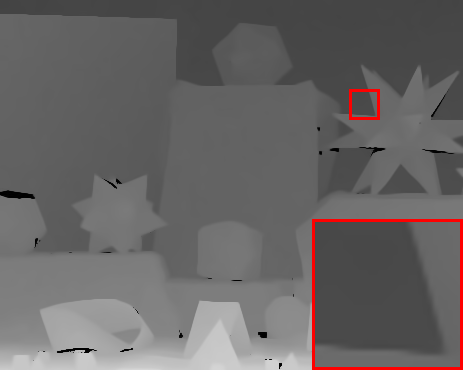}}
\subfigure[WNNM / 40.28dB]{\includegraphics[width=1.4in, angle=0]{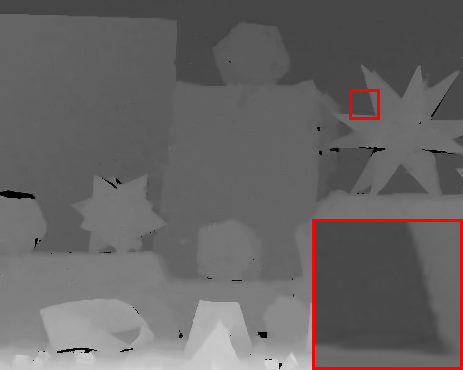}}
\subfigure[GNNLG / 41.37dB]{\includegraphics[width=1.4in, angle=0]{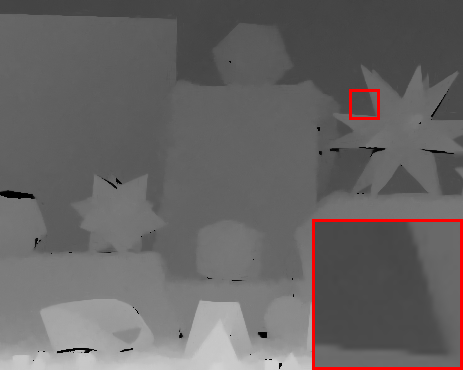}}
\caption{The visual comparison of depth image $Moebius$ when AWGN variable $\sigma$ = 15. }
 \end{figure*}

\begin{figure*}
\centering
\subfigure[Src image / PSNR]{\includegraphics[width=1.4in, angle=0]{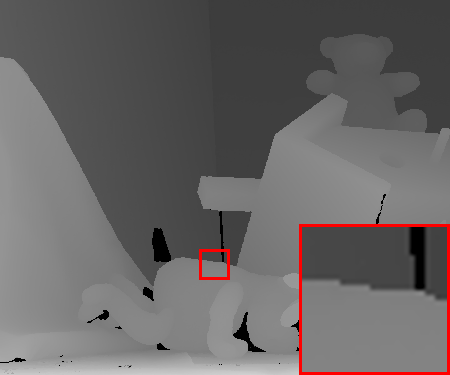}}
\subfigure[Noisy image / 22.14dB]{\includegraphics[width=1.4in, angle=0]{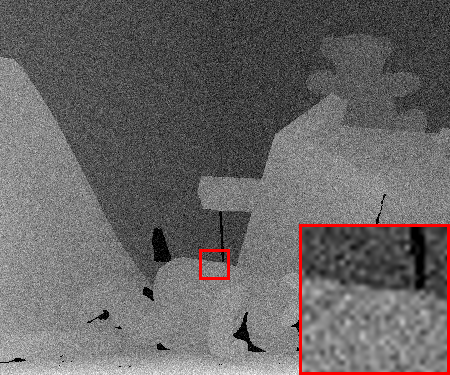}}
\subfigure[NLM / 36.67dB]{\includegraphics[width=1.4in, angle=0]{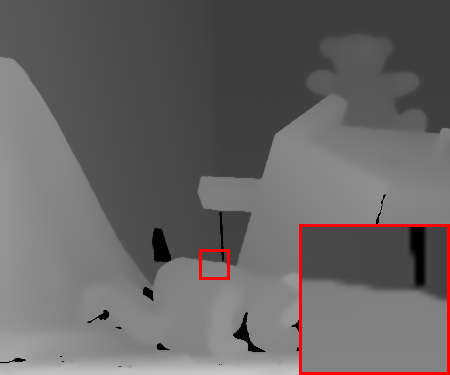}}
\subfigure[CSR / 35.86dB]{\includegraphics[width=1.4in, angle=0]{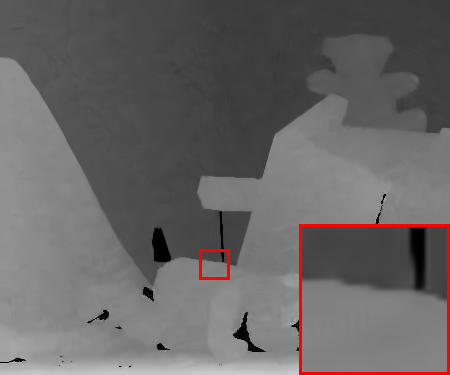}}
\subfigure[NLGBT / 36.98dB]{\includegraphics[width=1.4in, angle=0]{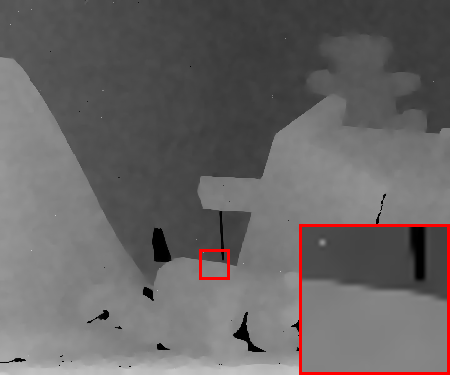}}
\subfigure[BM3D / 37.70dB)]{\includegraphics[width=1.4in, angle=0]{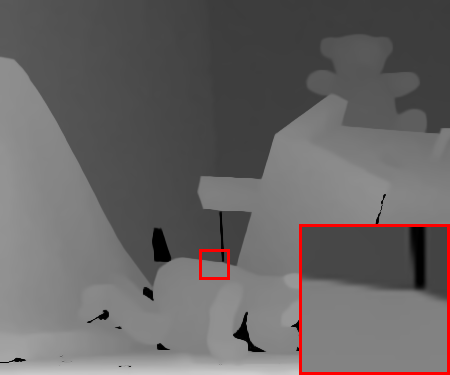}}
\subfigure[WNNM / 37.27dB]{\includegraphics[width=1.4in, angle=0]{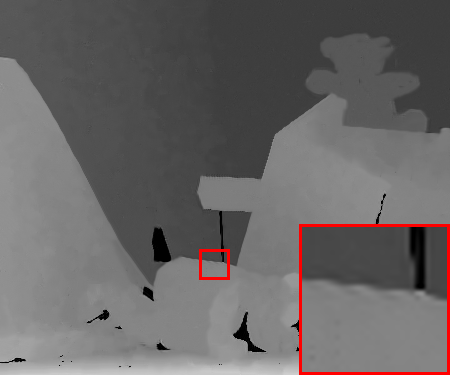}}
\subfigure[GNNLG / 38.00dB]{\includegraphics[width=1.4in, angle=0]{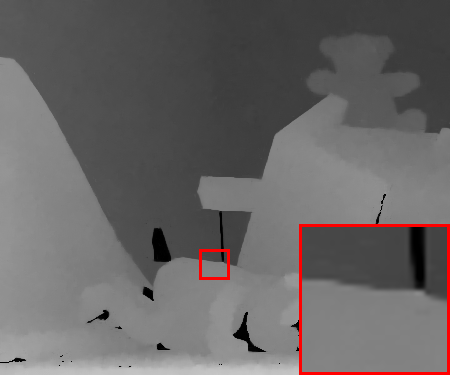}}
\caption{The visual comparison of depth image $Teddy$ when AWGN variable $\sigma$ = 20. }
 \end{figure*}

\begin{figure*}
\centering
\subfigure[Src image / PSNR]{\includegraphics[width=1.4in, angle=0]{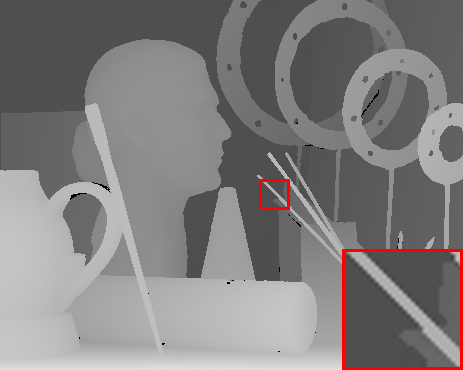}}
\subfigure[Noisy image / 20.18dB]{\includegraphics[width=1.4in, angle=0]{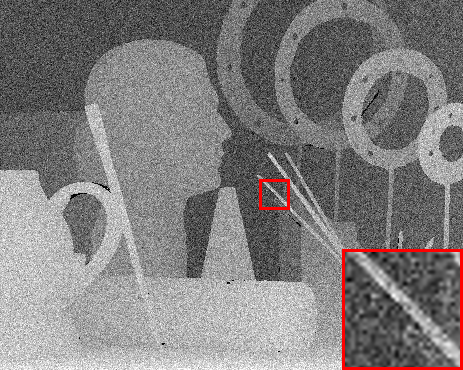}}
\subfigure[NLM / 35.23dB]{\includegraphics[width=1.4in, angle=0]{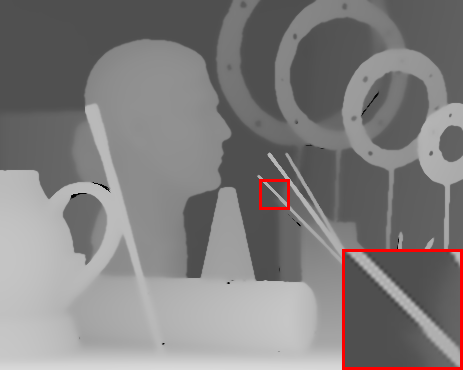}}
\subfigure[CSR / 34.23dB]{\includegraphics[width=1.4in, angle=0]{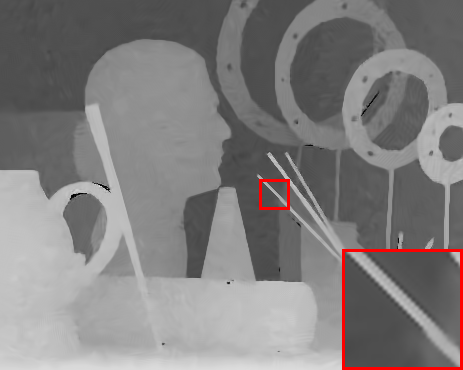}}
\subfigure[NLGBT / 34.34dB]{\includegraphics[width=1.4in, angle=0]{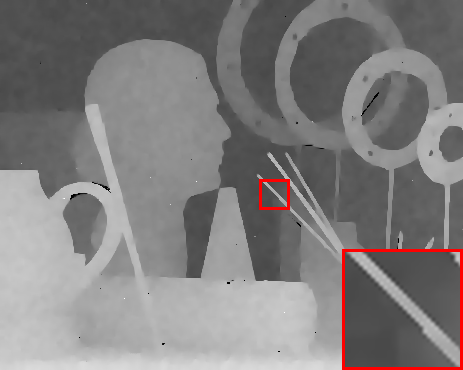}}
\subfigure[BM3D / 35.83dB)]{\includegraphics[width=1.4in, angle=0]{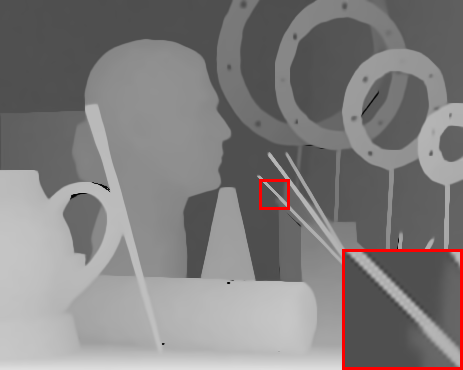}}
\subfigure[WNNM / 35.29dB]{\includegraphics[width=1.4in, angle=0]{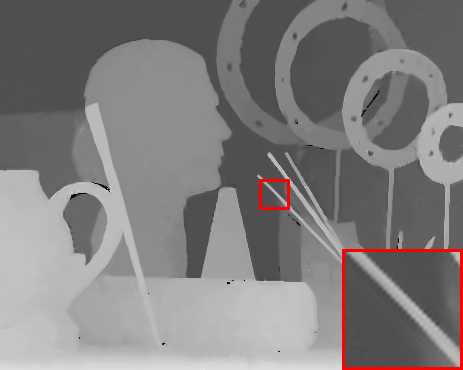}}
\subfigure[GNNLG / 36.32dB]{\includegraphics[width=1.4in, angle=0]{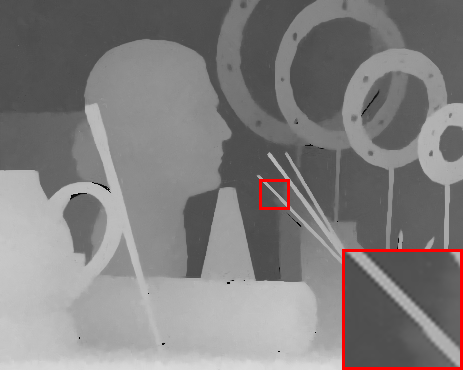}}
\caption{The visual comparison of depth image $Art$ when AWGN variable $\sigma$ = 25. }
 \end{figure*}

\begin{figure*}
\centering
\subfigure[Src image / PSNR]{\includegraphics[width=1.4in, angle=0]{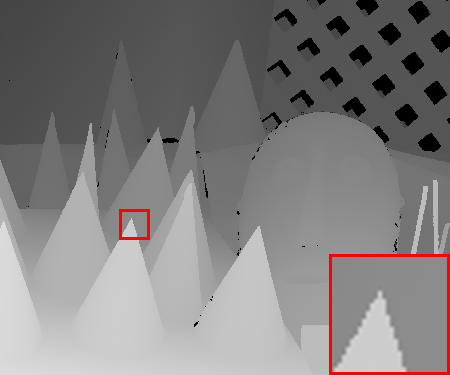}}
\subfigure[Noisy image / 18.58dB]{\includegraphics[width=1.4in, angle=0]{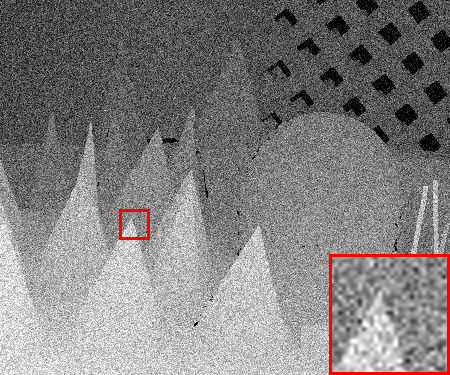}}
\subfigure[NLM / 32.84dB]{\includegraphics[width=1.4in, angle=0]{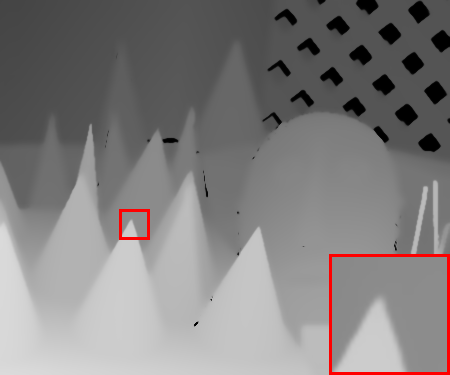}}
\subfigure[CSR / 32.55dB]{\includegraphics[width=1.4in, angle=0]{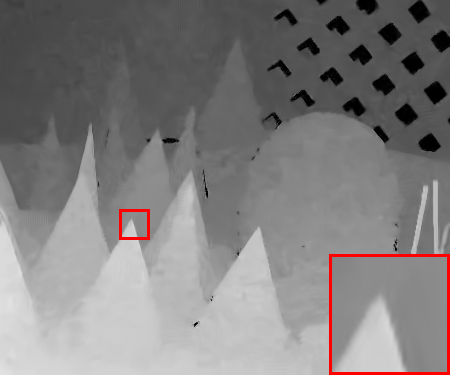}}
\subfigure[NLGBT / 32.84dB]{\includegraphics[width=1.4in, angle=0]{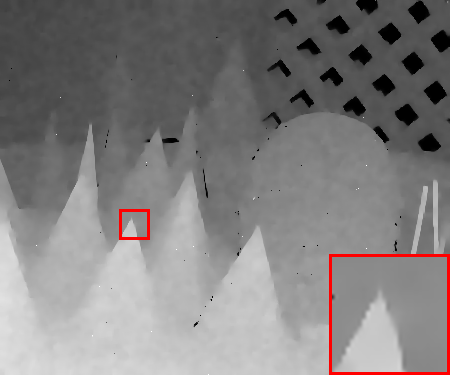}}
\subfigure[BM3D / 33.66dB]{\includegraphics[width=1.4in, angle=0]{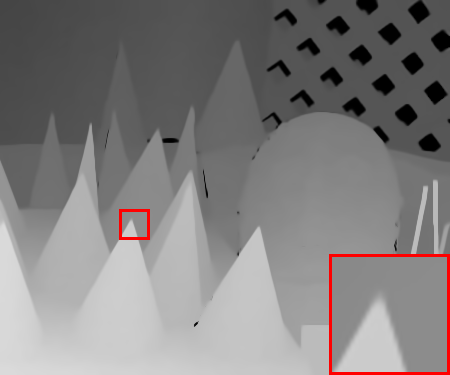}}
\subfigure[WNNM / 33.92dB]{\includegraphics[width=1.4in, angle=0]{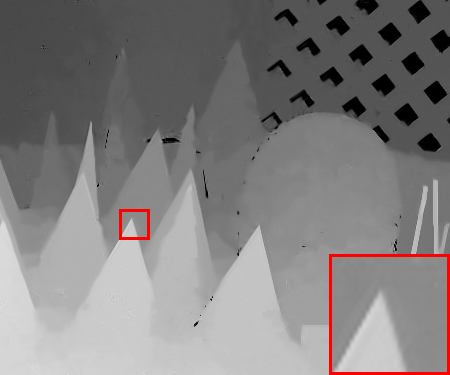}}
\subfigure[GNNLG / 34.72dB]{\includegraphics[width=1.4in, angle=0]{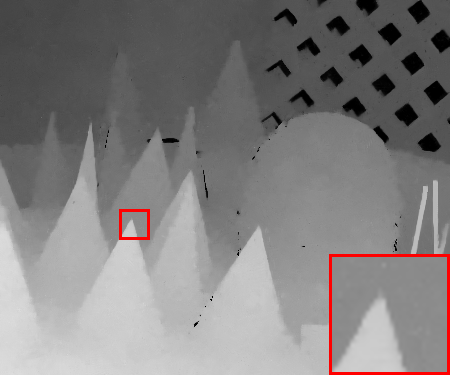}}
\caption{The visual comparison of depth image $Cones$ when AWGN variable $\sigma$ = 30. }
 \end{figure*}

\bibliographystyle{ACM-Reference-Format}
\bibliography{refer}
\end{document}